\pdfoutput=1
\documentclass{article}

\usepackage[preprint]{neurips_2026}

\usepackage[utf8]{inputenc}%
\usepackage[T1]{fontenc}%
\usepackage{hyperref}%
\usepackage{url}%
\usepackage{booktabs}%
\usepackage{amsfonts}%
\usepackage{nicefrac}%
\usepackage{microtype}%
\usepackage{xcolor}%
\usepackage{graphicx}
\usepackage{subcaption}

\usepackage{amsmath}
\usepackage{amsthm}
\usepackage{xspace}%
\theoremstyle{definition}
\newtheorem{definition}{Definition}[section]

\newcommand{\thq}[1][q]{\mathbf{th}_{#1}}

\newcommand{\task}{\texttt{task}}
\newcommand{\AI}{\texttt{AI}}
\newcommand{\tasks}{\textbf{tasks}}
\newcommand{\AIs}{\textbf{AIs}}

\title{On the estimation and validity of AI time horizons---a statistical look at the METR plot}

\author{%
  Drew T.~Nguyen \\
  Department of Statistics\\
  UC Berkeley\\
  \And
  William~Fithian \\
  Department of Statistics\\
  UC Berkeley\\
}

\begin{document}

\maketitle

\begin{abstract}
    METR's 50\% time horizon measures the human completion time of software tasks that an AI solves with 50\% probability, allowing AI capabilities to be expressed in interpretable units. On 228 tasks and 26 AIs, we recompute the time horizons using splines and item-response theory to relax the assumption that the AI difficulty of a task depends linearly on the log of human time. Our fitted spline can be interpreted as a function that \emph{converts} human time to AI difficulty; it is nearly flat in a region from 2--30 min but close to linear elsewhere. Hence, a time-horizon jump from 3 min to 30 min is much easier than one from 30 min to 5 hours despite the same multiplier of $10 \times$. Overall, we contribute time-horizon point estimates that perform better under a cross-validated suite of proper scoring rules, as well as diagnostic plots for assessing time horizons' construct validity. We suggest that time horizons be interpreted together with the diagnostic plots, especially as new time-horizon-based benchmarks are proposed or existing ones grow to include longer tasks. 
\end{abstract}

\section{Introduction}

How quickly are AI capabilities advancing? This fundamental question underlies the large bets and policy discussions regarding AI infrastructure today. Among researchers, opinions diverge wildly; high-profile pieces such as AI 2027 \citep{AI2027} and AI as Normal Technology \citep{AINormalTechnology} offer starkly contrasting views. However, the work of METR, a research non-profit in Berkeley, CA, stands out as a key source of guidance. Called ``the world's informal A.I. umpire'' by the \emph{New York Times} \citep{wittOpinionAIPrompt2025}, they provide compelling evidence that, in a certain sense, AI capabilities are advancing very quickly indeed. 

According to METR's ``time horizons'' methodology, each of $228$ hand-designed software engineering tasks is assigned a task length, or ``human time''---the length of time it takes a professional software engineer to successfully execute it. Then, crucially, each major AI release is assigned a measurement called its \emph{time horizon}---the human time of tasks that it can solve autonomously with 50\% probability, over repeated stochastic runs. An AI's time horizon hence captures its capability at solving software tasks in some precisely defined sense, linked to the suite of tasks.

For example, GPT-4's estimated time horizon is 4 minutes. This means, approximately, that on a software engineering task that takes a human 4 minutes to solve (such as ``Google this fact''), GPT-4 autonomously completes it with 50\% probability. The longer an AI's time horizon, the longer the tasks it can solve at this 50\% rate. 

\begin{figure}[tp]
    \centering
    \includegraphics[width=\textwidth]{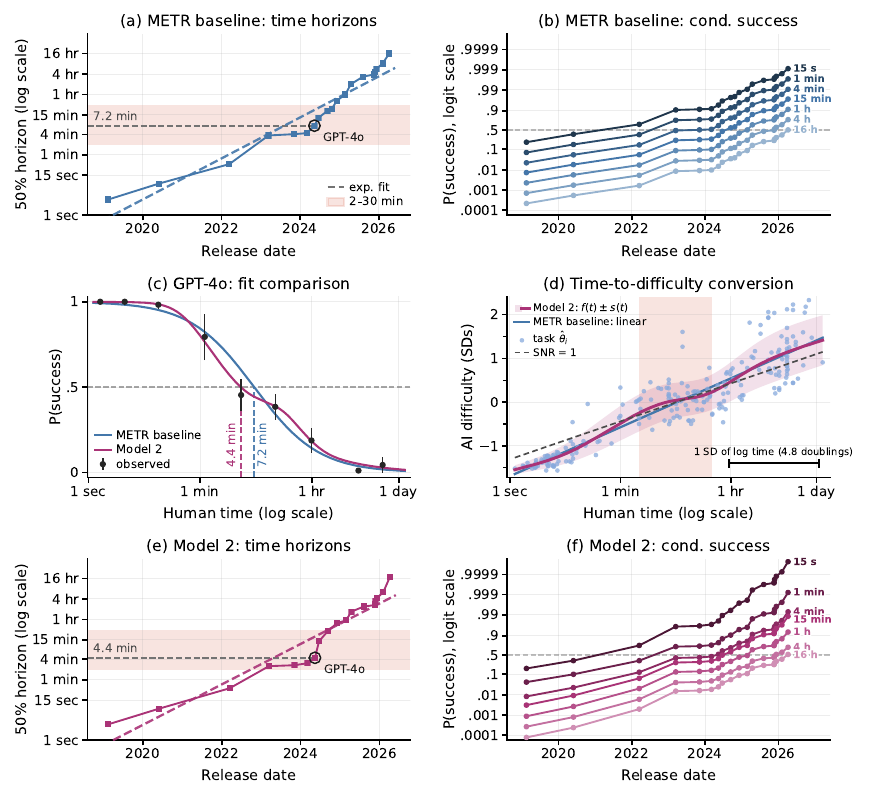}
    \captionsetup{singlelinecheck=false}
    \caption[Summary of contributions]{\textbf{Summary of contributions}. In this work we propose new point estimates for time horizons with an in-sample fit comparison (first column) as well as diagnostic charts for construct validity (second column). Details of how the first column was computed are described in Section \ref{sec:th-estimation}, including a description of Model 2, and details and interpretation for the second column are in Section \ref{sec:th-dh-cv}.
    \begin{itemize}
        \item[\textbf{(a)}] METR's time horizon chart, computed with the baseline shared-slope logistic method.
        \item[\textbf{(b)}] A conditional success plot of $\Pr(\text{success} \mid \text{human time})$ based on the baseline method, which is a more complete summary of the fit than 50\% time horizons. The time horizons can be read off by locating where the dotted line intersects with any plotted point, and inspecting that point's time label. 
        \item[\textbf{(c)}] An in-sample comparison of the fitted $\Pr(\text{success} \mid \text{human time})$ for GPT-4o. Model 2 is closer. 
        \item[\textbf{(d)}] The time-to-difficulty conversion plot, using Model 2's conversion function $f$ and fitted standard deviation $s$. Each of the 228 points represents a task, plotted with its human time and estimated AI difficulty. The AI difficulty scale is arbitrary up to affine reparametrization, so we plot in standardized units. On these tasks log human time has a good signal to noise ratio (SNR) for AI difficulty overall, while having a notable kink from 2--30 min. The baseline assumes linearity.
        \item[\textbf{(e)}] Time horizons under Model 2. The scoring rule-based evaluations in Section \ref{sec:fit-results} suggest that the estimates in the shaded region are particularly better; they are also the most different.
        \item[\textbf{(f)}] A conditional success plot based on Model 2. Beyond just estimating 50\% time horizon, scoring rule-based evaluations also suggest that Model 2 predicts success better overall compared to the baseline, indicating that (f) may better summarize the data compared to (b).
    \end{itemize}}
    \label{fig:headline}
\end{figure}

Using METR's public data, we reproduce\footnote{Strictly speaking, we use the shared-slope logistic approach described by \citet{ImpactModellingAssumptions2026} in a METR research note.} METR's \emph{time-horizon plot} in Figure \ref{fig:headline}(a). Originally released in March 2025, it was updated point by point with each new AI release until ending with Claude Mythos in May 2026. The plot depicts an exponential increase in the time horizons, convincing many of AI's overall exponential advance---the \emph{New York Times} compared it to Moore's law \citep{rooseHowYouMeasure2026}---and, presumably, convincing many of AI's future investment returns as well. Meanwhile, writing from the perspective of AI safety, \citet{BuildingTechnologyDrive2026} relates it to the Keeling Curve, noting that, where the original Keeling Curve demonstrated rapidly rising atmospheric CO$_2$ concentrations and oriented global strategy, the time-horizon plot could serve the same role for AI risk. Although the prospect of slowing down AI development became U.S. front page news in summer 2026, the time-horizon plot helped catalyze advocacy for a slowdown before then---including from figures such as Turing Award winner Yoshua Bengio and U.S. Senator Bernie Sanders. 

Unlike Moore's law or the Keeling Curve, however, the quantity plotted on the y-axis is not a direct measurement of a physical quantity, but rather a statistical estimate which can be critiqued and improved. Indeed, the time-horizon methodology can be viewed as an application of item-response theory (IRT) techniques with human time as a covariate, allowing ``software engineering capability'' to be measured in the highly interpretable units of human time. This move---anchoring capability in an external scale---is the time-horizon plot's key idea.

In this work, we critique the time-horizon plot by reanalyzing METR's data and fitting new models. Our methods and framing may be useful for any implementation of IRT methodology using an external anchoring scale. We answer two questions:
\begin{itemize}
    \item What is the METR plot trying to estimate, and can a better estimate be devised? (Section \ref{sec:th-estimation})
    \item How useful are time horizons for measuring AI capabilities? How can we diagnose this? (Section \ref{sec:th-dh-cv}) 
\end{itemize}
Though we are not the first to critique the METR plot, we are unaware of any critiques that fully claim a better estimator, or that discuss whether and under what conditions they are substantively useful. 

In Section \ref{sec:th-estimation}, we improve the estimation of the time horizons on held-out data. Our main proposal is two statistical models, Model 1 and Model 2, that lead to new estimates; we quantify their improvement via scoring rule criteria. An in-sample diagnostic showing the improved fit of Model 2 relative to the baseline is presented in Figure \ref{fig:headline}(c); its time horizon estimates are in Figure \ref{fig:headline}(e). 

Next, when and whether time horizons themselves are useful is a question of their \emph{construct validity}, and it depends crucially on the relationship between a task's human time and the latent difficulty of the task itself. Our Model 2, an item-response theory model, estimates this relationship. 

Therefore in Section \ref{sec:th-dh-cv}, using Model 2, we propose two plots to help diagnose the construct validity of the time horizons: a ``time-to-difficulty conversion'' plot and a ``conditional success trajectory'' plot. Here ``conditional'' means conditional on human time. 

Figure \ref{fig:headline}(d) shows the time-to-difficulty conversion plot, as well as the time-to-difficulty conversion function $f$. The plot displays estimates of each of the 228 tasks' AI difficulty, which are non-linear in log human time. In particular, we highlight a flat region between 2--30 minutes, which also can be seen in Figure \ref{fig:headline}(c); this both raises the question of just how meaningful a time horizon jump within that region is, while also suggesting that a jump cannot have the \emph{same} meaning there as it does elsewhere.

Figures \ref{fig:headline}(b) and (f) show our conditional success trajectory plots, though we believe that Figure \ref{fig:headline}(f) is a better fit to the data than (b). This figure suggests that a time horizon jump from 4 min to 15 min may be quite small indeed.

Finally, in Section \ref{sec:discussion} we discuss what this may mean for the time horizon methodology when potentially extended to longer tasks or applied to other domains. Overall, we view our work as providing a constructive critique for benchmarks like METR's. We improve the time horizon estimates, while also providing a lens to assist in their interpretation.

\paragraph{Note on terminology} The word ``model'' is used in AI contexts to refer to a large language model, and in applied statistics contexts to refer to a statistical model or fit. Here, we will reserve ``model'' for the latter, and refer to large language models as ``AIs''. 

\subsection{Prior work}\label{sec:prior-work}

Abstractly, item-response theory \citep{bakerBasicsItemResponse2001} is the body of statistical techniques suited for jointly learning respondent ability and item difficulty from outcomes observed for an item-respondent pair. METR's work as well as ours can be thought of as an application of this kind of theory. 

The original time-horizon measurement was introduced by \citet{kwaMeasuringAIAbility2025a}, but the basic idea of pinning a unitless measure of ability, from IRT, to an externally interpretable quantity has previously been proposed; see, for example, the Lexile Framework for reading \citep{stennerMeasuringReadingComprehension2023}. 

Since 2025, there have been many critiques of \citet{kwaMeasuringAIAbility2025a} on 
Substack, LessWrong, and the EA Forum \citep{shash42HowGameMETR2025, witkinMETRGraph2026,gregorylewisGeneralCapabilityCapabilities2026}.
Statistical critiques and extensions of the time-horizon plot include \citet{mossUpdatedMETRsData2026} and METR's own research notes \citep{ClarifyingLimitationsTime2026,ImpactModellingAssumptions2026}. Substantive construct validity critiques of the time horizon concept in domains outside of software tasks include \citet{HowDoesTime2025, mertensCrashingWavesVs2026a}. 

A major alternative measure of AI progress, similarly based on item-response theory, is the Epoch Capabilities Index \citep{hoRosettaStoneAI2025}. Unlike the time-horizon approach, they do not pin an AI's capability to an externally interpretable measure; this allows them to combine many benchmarks into one measure, at the cost of a less interpretable index. 

\section{The time-horizon estimation problem}\label{sec:th-estimation}

\subsection{The statistical framework}\label{sec:framework}

What problem is the time-horizon plot solving? Within the framework of a statistical model, it depicts estimates of a well-specified quantity, as we shall see in Definition \ref{def:timehorizon}. 

Suppose we are able to draw tasks $(\task_i)_{i = 1}^I$ from some task distribution $\mathcal{D}$, and fix AIs $(\AI_j)_{j = 1}^J$. We think of the tasks as units existing in some space $\tasks$, and AIs in $\AIs$. 

Each task is associated with certain covariates, in particular associated with a human time $T_1, \dots, T_I > 0$ and a task family ID. On the other hand, each AI is associated with a \emph{release date} $D_1 < \dots < D_J$. Let $\mathcal{P} : \tasks \times \AIs \to [0,1]$ be the function that associates each $(\task, \AI)$ pair with the probability that $\AI$ solves $\task$. Denote $p_{ij} := \mathcal{P}(\task_i, \AI_j)$. 

Given $(\task_i)_{i = 1}^I$ and $(\AI_j)_{j = 1}^J$, assume we have access to a mechanism to draw iid samples $Y_{ij} = (Y_{ij1}, \dots, Y_{ijn_{ij}})$, such that $\mathbb{E} Y_{ij1} = p_{ij}$, which models multiple trials of re-running the AI on the same prompt. Here $Y_{ij}$ represents the $n_{ij}$ runs where $\AI_j$ attempts $\task_i$, and $Y_{ijr} \in \{0,1\}$ is the success or failure of the $r$th run. 

(Despite this iid assumption, approximation errors in the functional form of any practical model specification will induce correlation among trials in the runset $Y_{ij}$ even for the same AI-task pair when conditioning on latent features. We return to this point when introducing our Model 2.)

Define the success curve $p_{j}(t) = \mathbb{E}_{\task_i \sim \mathcal{D}}[Y_{ij1} \mid T_i = t]$. We make the structural assumption that, for our chosen AIs and task distribution $\mathcal{D}$, this function is strictly monotone decreasing in $t$, so that its inverse is always defined.

Now, we can define the oracle quantity we would like to estimate, from the data $(Y_{ijr})$ and $(T_i)$.  

\begin{definition}\label{def:timehorizon}
    For $q \in (0,1)$, the \emph{$q$-time horizon} of $\AI_j$ is 
    \[
        \thq(\AI_j) := p_{j}^{-1}(q).
    \]
\end{definition}

This is the human time of tasks from $\mathcal{D}$ that $\AI_j$ can solve with probability $q$. To learn this quantity for each AI it suffices, therefore, to learn the regression function $p_j(t)$, for each $\AI_j$. The time-horizon plots of Figures \ref{fig:headline}(a) and (e) are then a scatter of points $(D_1, \widehat\thq(\AI_1)), \dots, (D_J, \widehat\thq(\AI_J))$. 

Importantly for the interpretation of the METR plot, the release dates $D_1, \dots, D_J$ are used nowhere in the model; the observed exponential growth is entirely empirical and we do not challenge it here. 

\subsection{Implementing the framework}\label{sec:implementation}

In order to implement this framework in the real world, we require the following ingredients: tasks $(\task_i)_{i = 1}^I$, thought to be sampled from some relevant task distribution $\mathcal{D}$; a setup for humans and AIs to attempt the tasks, and hence measure each task's human time, $(T_i)_{i=1}^I$, as well as the task-AI success indicators $Y_{ij}$ for each task-AI pair $(i,j)$; a technique for estimating the marginal success probability $p_j(t)$ for the $j$th AI, with confidence intervals; and finally, a method for evaluating whether our estimate had good performance. 

In this section, we summarize strategies for each of these implementation components. In particular, we summarize METR's approach for the first two components; see \citet{kwaMeasuringAIAbility2025a} for a fuller description. For the last two components (estimation and evaluation), we summarize general principles here and defer fuller exposition to Sections \ref{sec:modelspecs} and \ref{sec:evaluation}. 

\paragraph{Designing the tasks.} METR designed 228 tasks in their Time Horizon 1.1 framework, which were broadly software engineering relevant tasks such as ``implement this function'' or ``train this classifier'' written as a set of instructions, together with an automated function that scores the result. Each task was a member of one of three ``task suites'' with a unifying theme; each task was further classified into a \texttt{task\_family}, of which there were 79. 

\paragraph{Attempting the tasks.} For each task, a shared setup was used for both humans and AIs to attempt the task. This was concretely three things per task: a set of task instructions, a field to input the task solution, and a scorer function to judge success or failure. 

On the human side, multiple software engineers were recruited to attempt each task while being timed (typically ${\sim}4$ attempts per task); on attempts where the scorer judged a success, their time measurements were combined using the geometric mean. This value is called the \emph{human time} of the task (METR's ``task length''), which we denote $T_i$ for $\task_i$. However, these measurements were unobtainable or deemed invalid (e.g. too few successes) for 29\% of the tasks, and in those cases, an expert's value is used instead to set as $T_i$. 

On the AI side, the $j$th AI was provided the instructions for $\task_i$ as a prompt and run autonomously $n_{ij}$ times, and the scorer's output recorded as $Y_{ij} = (Y_{ij1}, \dots, Y_{ijn_{ij}})$ for each run. Note that we ignore the \emph{time AIs take} to complete tasks.

\paragraph{Fitting models and estimating time horizons.} Here we summarize general principles for fitting models; concrete approaches are detailed in Section \ref{sec:modelspecs}. The framework of Section \ref{sec:framework} allows us to precisely define the statistical problem at hand: we want good estimates for $\thq$, especially $\thq[0.5](\AI_j)$ and $\thq[0.8](\AI_j)$, for each $\AI_j$. METR's original approach, as well as ours, will be to first obtain estimates for $p_j(t)$ and then invert the function according to Definition \ref{def:timehorizon}. These per-AI success curves $p_j(t)$, also called ``characteristic curves'' in item-response theory, can be estimated in any way in principle, as long as they are ultimately evaluated with held-out data. 

These approaches are all based on the same tabular dataset, where a row consists of the AI's ID $j$, the task's index $i$, the run index $r$, the success indicator $Y_{ijr}$ for the AI's attempt on that task, and finally the task's human time $T_i$---we do not use the AI's release date $D_j$ for fitting, but only for plotting.

\paragraph{Evaluating models.} Finally, we summarize general principles for evaluating models; our evaluation metrics are defined in Section \ref{sec:evaluation}, with results in Section \ref{sec:fit-results}. Essentially, given estimates for the success curves $p_1(\cdot), \dots, p_J(\cdot)$, we compute these metrics to adjudicate between the different estimates and select the best. Our metrics consist of proper scoring rules and Murphy diagrams \citep{dimitriadisEvaluatingProbabilisticClassifiers2023}, which we computed with 5-fold cross validation; this can be thought of as an extension of the evaluations in \citet{ImpactModellingAssumptions2026}. We also plot in-sample visual diagnostics for our $p_j(t)$ estimates against empirical frequencies to assess the functional form, as done by \citet{kwaMeasuringAIAbility2026} in their Figure 4. 

\subsection{Model specifications} \label{sec:modelspecs}

In this section, we list the four main model specifications we ran. By model specification, we mean a model family and fitting procedure for the success curves $p_j(t)$. The first two were proposed by METR, and the last two are our contribution. 

\paragraph{Logistic regression.} This is the original headline model specification used by METR. For each $\AI_j$, we assume for $p_j(t)$ the log-linear model
\[
    \log\left(
      \frac{p_j(t)}{1-p_j(t)}
    \right)  = \alpha_j - \beta_j \log_2(t)
\]
with $\beta_j > 0$. We estimate the parameters $\alpha_j$, $\beta_j$ for each $j$ with logistic regression, i.e. minimizing binary cross entropy, using $\log_2(T_1), \dots,\log_2(T_I)$ as covariates. When the cross-entropy is unweighted, this is formally equivalent to maximum likelihood with the generative model $Y_{ijr} \mid T_i = t \sim \operatorname{Bern}(p_j(t))$ independently for all $i, j, r$. 

Such an independence assumption is certainly wrong, but this does not necessarily mean that the success curves are wrong. Also, tasks in a shared task family have correlated behavior, so following \citet{kwaMeasuringAIAbility2025a}, the effect of large families was downweighted in training by using the sqrt-family weighting described in Section \ref{sec:weights} on the cross-entropy loss.

\paragraph{Logistic regression with shared slope (our baseline).} This specification was tried by \citet{ImpactModellingAssumptions2026} in work for METR. It is the same as the previous, except we assume for $p_j(t)$ that
\[
    \log\left(
      \frac{p_j(t)}{1-p_j(t)}
    \right)  = \alpha_j - \beta \log_2(t)
\]
for a fixed $\beta > 0$ common to each AI. Again, an unweighted fit is equivalent to maximum likelihood on a model where all the $Y_{ijr}$ are independent conditionally on the human times $T_1, \dots, T_I$. 

\citet{ImpactModellingAssumptions2026} observes it has better performance in cross-validation on both Brier score and marginal log score, with sqrt-family weighting; we replicate this finding, and in fact it beats METR's model spec on all but one of the 12 metrics in our proper scoring suite (Section~\ref{sec:fit-results}). We hence use this spec as our performance baseline, rather than the logistic regression spec. The time horizons from this fit were displayed in Figure~\ref{fig:headline}(a). 

\paragraph{Model 1: Shared monotone spline, with shared slope.} This specification just replaces the linear dependence on $\log_2(t)$ with a monotone function. We assume for $p_j(t)$ that 
\[
    \log\left(
      \frac{p_j(t)}{1-p_j(t)}
    \right) = \alpha_j - \beta\, f(\log_2 t), \qquad \beta > 0.
\]
We take $f$ to be a monotone natural cubic spline; specifically, a monotone natural cubic I-spline with nonnegative coefficients with 4 interior knots. Additionally, because of non-uniqueness with respect to affine transformations of $f(\cdot) \mapsto a f(\cdot) + b$ where $a > 0$, we make the identifiability assumption $f(\min \log_2 T_i) = \min \log_2 T_i$ and $f(\max \log_2 T_i) = \max \log_2 T_i$. Then we fit with binary cross entropy, with sqrt-family weighting; again, an unweighted fit is equivalent to maximum likelihood on a model where all the $Y_{ijr}$ are independent conditionally on the human times $T_1, \dots, T_I$.

A monotone spline-based approach was previously considered by METR in \citet{ClarifyingLimitationsTime2026}, and attempted in \citet{ImpactModellingAssumptions2026}, but did not beat the baseline; this was likely because they were \emph{separate} splines, rather than shared. 

The single shared shape $f$ gives this specification its interpretation. It is an estimate of how AI difficulty varies with human time, common to all the AIs. Using item-response theory in Model 2, we can extend this interpretation further.

\paragraph{Model 2: Explanatory IRT with family effects and overdispersion.} This specification is a fully generative model $\mathcal{M}$ for all the $Y_{ijr}$'s. We fit it with maximum likelihood, and subsequently derive the success curves. This model involves task families; let $F$ index the families, and let $F(i)$ be the task family of $\task_i$. Let $\check T_{{F(i)}}$ be the geometric mean of human times in $F(i)$. 

In this model, each $\task_i$ carries a latent difficulty $\theta_i \in \mathbb{R}$. Define
\[
    \tilde p_{ij}(\theta_i) := \Pr(Y_{ijr} = 1 \mid \theta_i)
\]
for the success probability of a single run conditional on the difficulties. We model this as
\[
    \log\left(
      \frac{\tilde p_{ij}(\theta_i)}{1 - \tilde p_{ij}(\theta_i)}
    \right) = \alpha_j - \beta \theta_i,
\]
where the $\beta > 0$ is shared among all AIs $j$. The latent difficulty is modeled as
\[
    \theta_i = f(\log_2 T_i) + \xi_{F(i)} + \varepsilon_i,
    \qquad
    \xi_F \sim \mathcal{N}\big(0,\ \tau(\log_2 \check T_F)^2\big),
    \qquad
    \varepsilon_i \sim \mathcal{N}\big(0,\ \sigma(\log_2 T_i)^2\big),
\]
for some mean function $f$, family variance $\tau(\cdot)^2$, and idiosyncratic variance $\sigma(\cdot)^2$. 

Even though we assume $Y_{ij1}, \dots, Y_{ijn_{ij}}$ are iid, we do not model them as independent conditional on $\theta$, since the pair $(\theta_i, \alpha_j)$ is not a complete summary of the AI-task interaction, and conditioning on these induces correlation between trials for each AI-task pair. Hence we introduce a correlation parameter $\rho \in (0,1)$ shared by all AIs, put $\varphi = (1-\rho)/\rho$, and model the final probability $p_{ij}$ as drawn from
\[
    p_{ij} \mid \theta_i \sim \operatorname{Beta}\big(\tilde p_{ij}(\theta_i)\,\varphi,\ (1 - \tilde p_{ij}(\theta_i))\,\varphi\big),
    \qquad Y_{ij1}, \dots, Y_{ijn_{ij}} \mid p_{ij} \overset{\text{iid}}{\sim} \operatorname{Bern}(p_{ij}),
\]
so that $\sum_{r = 1}^{n_{ij}} Y_{ijr}$ is beta-binomial, allowing us to model it with higher variance than if it were only binomial (hence the name ``overdispersion''). The source of this higher variance is the correlation $\operatorname{Corr}(Y_{ijr}, Y_{ijr'} \mid \theta_i) = \rho$ for $r \neq r'$. 

Fitting maximum likelihood under this model is not trivial. We implemented an EM algorithm and approximated the integrals with Gauss-Hermite quadrature. Note that, similar to Model 1, the likelihood is invariant to affine transformations of $\theta$, so we fix the scale with identifiability assumptions on $f$. 

To compute $p_j(t)$ under this model, we must marginalize over the random effects: 
\begin{equation}\label{eq:marginal-p}
    p_j(t) = \mathbb{E}\left[\left(1 + \exp(-(\alpha_j - \beta\theta))\right)^{-1}\right],
    \qquad \theta \sim \mathcal{N}\!\left(f(\log_2 t),\ \sigma(\log_2 t)^2 + \tau(\log_2 t)^2\right),
\end{equation}
where evaluating $\tau$ at $\log_2 t$ is the convention for a fresh task from a fresh family whose mean human time is $t$. We evaluate this by quadrature and invert numerically for $\widehat\thq$. 

\paragraph{Remarks on Model 2.} This is an item-response theory model. Specifically, it is a kind of 1PL Rasch model due to the shared $\beta$. In general, item-response theory uses latent variables to model the success probability of a respondent on a test item. It captures correlations between responses that share a respondent or an item by modeling a shared latent trait that connects them---a student's ability in a standardized testing setting, or, in our case, the task difficulty $\theta_i$, which connects $Y_{ij}$ and $Y_{ij'}$ for $j \neq j'$. 

It is an \emph{explanatory IRT} here because $\theta_i$ is regressed on the task's observed covariates: human time and task family, which ``explain'' the values of $\theta$. We use the same family of monotone spline $f$ as in Model 1, with the same identifiability assumption, and take $\sigma(\cdot)$ and $\tau(\cdot)$ to each be the exponential of a natural cubic spline with 2 interior knots. 

Like in Model 1, the mean function $f$ accounts for non-linearities between difficulty and log human time, while the individual and family effects account for possible spread in task difficulties from their mean. In particular, the family effect $\xi_F$ accounts for correlations among tasks within a family, without the heuristic sqrt-family reweighting. 

We account for dependence within the runset $Y_{ij1}, \dots, Y_{ijn_{ij}}$ via the correlation parameter $\rho$. In the data, this dependence is actually substantial: assuming the generative model implied by the baseline spec, where all $Y_{ijr}$ are independent, then assuming the fitted $p_{ij} := p_j(T_i)$ were correct, we would predict that $\sim 60\%$ of vectors $Y_{ij} = (Y_{ij1}, \dots, Y_{ijn_{ij}})$ were unanimous. However, it is actually 83\% of them. The correlation parameter affects the final fit by downweighting AI-task pairs with many runs, as they carry redundant information. 

\subsection{Evaluation with proper scoring rules}\label{sec:evaluation}

In this section, let $\mathcal{M}$ index modeling approaches, each of which supplies a success curve $p_{j, \mathcal{M}}(t)$ for every $\AI_j$, which we assume to be monotone decreasing in $t$, so that the time horizons $\thq^{\mathcal{M}}(\AI_j) := p_{j, \mathcal{M}}^{-1}(q)$ are well defined. Let
\[
    \mathbf{Y} = (Y_{ijr})_{i \leq I,\ j \leq J,\ r \leq n_{ij}},
    \qquad
    \mathbf{T} = (T_1, \dots, T_I)
\]
collect, respectively, every run outcome in the table and every human time, so that the dataset is the pair $(\mathbf{T}, \mathbf{Y})$. 

A scoring rule is then a function $S((\mathbf{T}, \mathbf{Y}), \mathcal{M})$ of the data and the approach. We use the four scores below. The first three (marginal log score, Brier score, and elementary score) are oriented so that smaller is better, and we report each divided by its total weight $\sum_{i,j} n_{ij} w_{ij}$.

\paragraph{Marginal log score.} Also known as binary cross entropy. Writing $\hat p_{ij} := p_{j, \mathcal{M}}(T_i)$ for the predicted success probability, let
\[
    S_{\mathrm{MLS}}((\mathbf{T}, \mathbf{Y}), \mathcal{M}) = \sum_{i = 1}^I \sum_{j = 1}^J w_{ij} \sum_{r = 1}^{n_{ij}} \left( -\,Y_{ijr} \log \hat p_{ij} \;-\; (1-Y_{ijr}) \log(1-\hat p_{ij})\right),
\]
for some weights $w_{ij} > 0$. This score only depends on $\mathcal{M}$ through the functions $p_{j, \mathcal{M}}(t)$. 

With all the weights equal to 1, minimizing this score in training is identical to maximizing log-likelihood when every $Y_{ijr}$ is assumed drawn independently. Even though such a model is certainly not a good fit in our application, that does not necessarily imply that the resulting success-curve functions $p_{j, \mathcal{M}}(T_i)$ are not a good fit. 

\paragraph{Brier score.} With the same definitions as in marginal log score, let
\[
    S_{\mathrm{BR}}((\mathbf{T}, \mathbf{Y}), \mathcal{M}) = \sum_{i = 1}^I \sum_{j = 1}^J w_{ij} \left( \sum_{r = 1}^{n_{ij}} \left(Y_{ijr} - \hat p_{ij}\right)^2 \right).
\]
Like the marginal log score, this only depends on $\mathcal{M}$ through the functions $p_{j, \mathcal{M}}(t)$. 

\paragraph{Elementary binary score at level $q$.} For $q \in (0,1)$, define
\begin{align*}
    S_{q}((\mathbf{T}, \mathbf{Y}), \mathcal{M}) 
        &= \sum_{i = 1}^I \sum_{j = 1}^J w_{ij} \sum_{r = 1}^{n_{ij}} \left( q (1 - Y_{ijr}) 1\{\hat p_{ij} > q\} + (1 - q) Y_{ijr} 1\{\hat p_{ij} < q\} \right) \\
        &= \sum_{i = 1}^I \sum_{j = 1}^J w_{ij} \sum_{r = 1}^{n_{ij}} \Big( q (1 - Y_{ijr}) 1\{T_i < \thq^\mathcal{M}(\AI_j)\} \\
        &\hphantom{{}= \sum_{i = 1}^I \sum_{j = 1}^J w_{ij} \sum_{r = 1}^{n_{ij}} \Big(}
            + \, (1 - q) Y_{ijr} 1\{T_i > \thq^\mathcal{M}(\AI_j)\} \Big)
\end{align*}
assuming no ties between $\hat p_{ij}$ and $q$. This score only depends on $\mathcal{M}$ through the time horizons $\thq$. In other words, this score directly measures the quality of the time horizon estimates. 

These scores are elementary in the sense that every proper scoring rule for a binary outcome is a nonnegative mixture of them \citep{schervishGeneralMethodComparing1989}; in particular, the Brier score satisfies $S_{\mathrm{BR}} = 2\int S_q\, dq$. The curve $q \mapsto S_{q}((\mathbf{T}, \mathbf{Y}), \mathcal{M}) $ is the Murphy diagram of $\mathcal{M}$ \citep{ehmQuantilesExpectilesConsistent2016}, which we also plot; one specification dominating another at every $q$ on the plot implies dominance under every proper score. 

At a single $q$, these scores are noisy, so for our evaluations we report the smoothed score $\tilde S_q((\mathbf{T}, \mathbf{Y}), \mathcal{M}) := \frac{1}{0.1}\int_{-0.05}^{0.05} S_{q + \varepsilon}((\mathbf{T}, \mathbf{Y}), \mathcal{M})\,  d\varepsilon$. This can be thought of then as measuring an average of the quality of the time horizons $\thq[q + \varepsilon]$ within that window. 

\paragraph{Full log-likelihood.} When a generative model is available, we can use this score. The (full) log-likelihood has the typical definition of
\[
    S_{\mathrm{JLL}}((\mathbf{T}, \mathbf{Y}), \mathcal{M}) := \log \Pr_\mathcal{M}(\mathbf{Y} \mid \mathbf{T}).
\]
This scores the entire model $\mathcal{M}$, including correlations between the $Y$. In particular, it is possible to do poorly in estimating the $p_j(t)$ but still do well on the full log-likelihood.

\subsubsection{Evaluation weights}\label{sec:weights}

The choice of weights $w_{ij} > 0$ essentially is a choice of what distribution of prediction problems we would like to average over; it does not depend on the run index $r$.

Writing $F(i)$ for the \texttt{task\_family} of $\task_i$ and $n_F$ for the number of scored tasks belonging to family $F$, we use the following three weightings:
\[
    w_{ij} = 1/n_{ij} \ \ (\text{equal-task}), \qquad
    1/(n_{F(i)}\, n_{ij}) \ \ (\text{equal-family}), \qquad
    1/(\sqrt{n_{F(i)}}\, n_{ij}) \ \ (\text{sqrt-family}).
\]
All of these schemes weight the AIs uniformly, but not the tasks. The equal task weighting targets a uniformly drawn task, adjusting for unequal numbers of runs. The equal family weighting instead targets a uniformly drawn task family, adjusting for unequal numbers of tasks in the family. Finally, the sqrt-family weighting is an intermediate used by METR for both fitting and evaluating their models. 

It is also possible to define versions of the scoring rules specific to each $\AI_j$, by restricting the sum over $j$ to that single AI. 

\subsubsection{A suite of proper scoring rule metrics}

Our main numerical measure of model quality is based on crossing the above scoring rules with the weighting schemes, under 5-fold cross validation on task families. That is, we split the 79 task families into 5 approximately equally sized folds, and then evaluate marginal log score, Brier score, and the smoothed elementary binary score at $q = 0.5$ and $q = 0.8$ on each fold using a model trained from data out-of-fold. Crossing these four scores with the three weighting schemes gives 12 proper scoring rule metrics. 

\subsection{Fit results and diagnostics}\label{sec:fit-results}

\begin{figure}[ht]
    \centering
    \includegraphics[width=\textwidth]{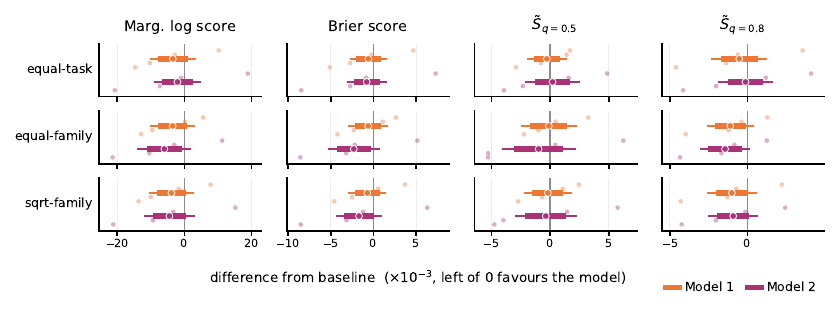}
    \caption{\textbf{Models 1 and 2 against the baseline across the scoring-rule suite.} Each panel shows the paired difference in weighted per-run score between our model and the baseline. (Left is better for our models.) The confidence intervals use family as the unit $n = 79$, at two nested levels: 80\% thick, 95\% thin. Point estimates for both of our models dominate the baseline at every displayed metric and weighting.}
    \label{fig:scoregrid}
\end{figure}

The results of the scoring rule metrics are displayed in Figure \ref{fig:scoregrid}, showing that the point estimates for our methods dominate the baseline on all metrics. Our chosen baseline itself dominates the original logistic regression specification on all metrics except one, namely equal-task $\tilde S_{q = 0.5}$, so we do not report its results.

\begin{figure}[th]
    \centering
    \includegraphics[width=\textwidth]{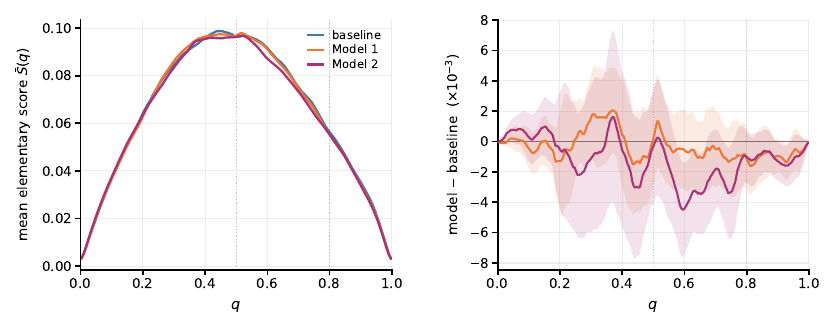}
    \caption{\textbf{Murphy diagrams of Models 1 and 2 against the baseline}, with equal-family weighting. \emph{Left:} Murphy diagrams of the baseline, Model 1, and Model 2. \emph{Right:} Each model's paired difference from the baseline ($\times 10^{-3}$), with its 95\% band based on family $n = 79$ units. For display, these curves are smoothed by a centered rolling mean over the $q$-grid ($\approx 0.03$ wide in $q$).}
    \label{fig:murphy}
\end{figure}

We also plot the Murphy diagrams in Figure \ref{fig:murphy}. For $q > 0.4$, the baseline seems to be nearly dominated, partly explaining why we get favorable performance on Brier score and marginal log score, as they are just weighted mixtures of points on the Murphy diagram. 

\begin{figure}[ht]
    \centering
    \includegraphics[width=\textwidth]{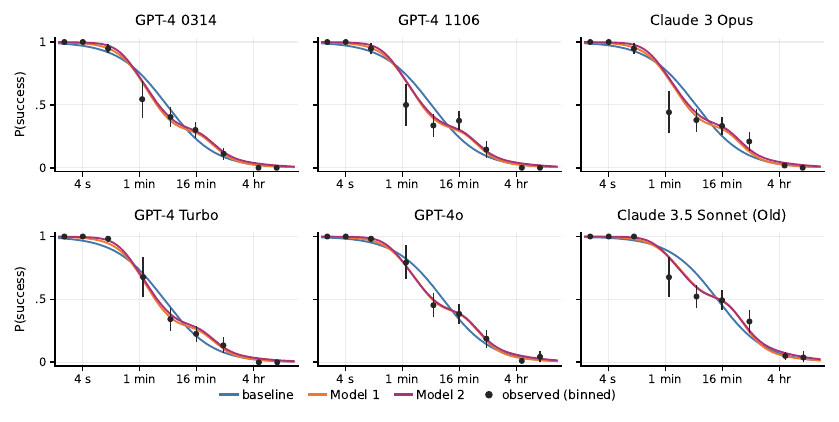}
    \caption{\textbf{In-sample fitted $p_j(t)$ against binned observed frequencies}, for six AIs with time horizons of 2--15 min. Black points pool each AI's runs within factor-4 human-time bins---width two in $\log_2$ minutes, with bin centers at 1 min, 4 min, 16 min, etc---plotted at the run-weighted mean human time within the bin. (The points are therefore unevenly spaced). Vertical bars are 95\% Wald intervals $\hat p \pm 1.96\sqrt{\hat p(1-\hat p)/N}$ over the bin's $N$ runs.}
    \label{fig:pfits}
\end{figure}

Additionally, we plot in-sample diagnostics in Figure \ref{fig:pfits} for six AIs with time horizons of 2--15 min. The shape of Models 1 and 2 bends to follow a flat region through roughly 2--30 minutes whose tasks Figure \ref{fig:headline}(d) shows have roughly equal difficulty, while the log-linear baseline is forced to cut straight through. Our fits substantially change the 50\% time horizon estimates for these AIs, some of which go outside the original METR plot's 95\% confidence intervals. 

A version separated by AIs is shown in the Appendix as  Figure~\ref{fig:pfits-all}, showing that a large amount of the gain in the metrics was due to improvement in estimates for AIs with time horizons of 2--15 min, such as these six, which were most affected by the flat region of tasks with human time 2--30 min, as shown in Figure \ref{fig:headline}(d).

\section{Diagnostics for construct validity}\label{sec:th-dh-cv}

\subsection{Construct validity for capability anchoring}

We loosely define \emph{construct validity} to be a property of a statistical estimand---whether, at the population level, it measures the \emph{construct} that we wish it to measure. For example, when reading the METR plot one might assume, implicitly, that the \emph{construct} of ``AI software development capabilities'' is well captured by the \emph{estimand} of ``50\% time horizons on METR's task suite''. But whether this is actually the case depends on how the human time of the suite's tasks relates to their difficulty for an AI---the relationship between human time and AI difficulty. 

This consideration applies to using any interpretable, external feature besides human time. Let us refer to this key idea of time horizons as ``capability anchoring''. Capability anchoring is in principle applicable to any benchmark dataset where $J$ AIs are set up to attempt $I$ tasks, or perhaps only a subset of them, and where the $I$ tasks have external annotations $T_1, \dots, T_I$. The general approach is then roughly abstractable to the following steps: 
\begin{enumerate}
    \item Learn task difficulties $\theta_1, \dots, \theta_I$ and AI capabilities $\alpha_1, \dots, \alpha_J$  jointly with some method (say, IRT), and hence learn a function $\tilde p(\theta_i, \alpha_j)$ as our best predictor of the chance that AI $j$ will succeed at task $i$.
    \item Regress the learned task difficulties $\theta_1, \dots, \theta_I$ on the external annotations $T_1, \dots, T_I$ to learn a mapping $f$ from annotation to difficulty. 
\item Define AI $j$'s ``anchored capability'' as a 50\% horizon, via $\thq[0.5] := f^{-1}(\theta_j^\star)$, where $\theta_j^\star$ solves $\tilde p(\theta_j^\star, \alpha_j) = 1/2$. (This is the annotation value whose predicted difficulty gives AI $j$ an even chance of success.) Use the ``anchored capability'' as an interpretable scale for the AI's capability. 
\end{enumerate}
In our setting, it is possible actually to follow these steps by using a Rasch model to perform Step 1 and a monotone spline to fit $f$ in Step 2. The function $p_j(t) := \tilde p(f(t), \alpha_j)$ then essentially follows the specification of Model 1 but with a different way of learning the parameters; we found the result to be nearly competitive with both Model 1 and 2. However, we prefer to use these steps as a conceptual way to understand methods for capability anchoring: Model 1 and the baseline implicitly perform Steps 1 and 2 jointly, though the baseline assumes that the mapping $f$ is linear in log human time. Model 2's EM algorithm alternates between Step 1 and Step 2 in the course of fitting, using the fitted $f$ to improve the estimates of difficulties and capabilities and then using those estimates to re-fit $f$. 

Given this abstraction for capability anchoring, we posit that the construct validity of time horizons and capability anchoring in general comes from affirmative answers to the following ``checklist'' questions: 
\begin{enumerate}
    \item Are the tasks representative of real tasks---does solving them mean the AI is ``capable'' in the sense that we wish?
    \item Do two numbers $(\theta, \alpha)$ and the function $\tilde p$ model the difficulty-capability interaction well, or do more numbers/different specifications fit significantly better? 
    \item Does the annotation have an objective meaning? Is it the same objective meaning across different tasks? Does this objective meaning capture a desired construct? Relative to this objective meaning, is the annotation measured well, and if not, can we account for it?
    \item (Predictiveness) Are the task annotations predictive of the difficulties, and hence AI success of the tasks?
    \item (Comparability) Is the relationship between the annotation and the difficulties different depending on the annotation? For anchored capabilities, can we interpret equal differences among AIs equally? If not, can we account for it?
\end{enumerate}

Many critiques of the construct validity of time horizons address the first three questions. In this work, we simply assume that answers to the first three questions are affirmative. Our diagnostic plots are then comparatively narrow---only addressing the last two questions of predictiveness and comparability. 

The issue raised by predictiveness is this. In the setting of Section \ref{sec:framework}, suppose AIs A and B have 50\% time horizons $T_A$ and $T_B$, with $T_B \gg T_A$, representing a large jump in the time horizon. Interpreting this as a large increase in capabilities seems misleading if both $p_B(T_A)-p_A(T_A)$ and $p_B(T_B)-p_A(T_B)$ are small, i.e. if human time is not predictive of AI success in $[T_A, T_B]$. This would be reflected by $p_B(\cdot)$ and $p_A(\cdot)$ being flat in that region. 

For comparability, suppose AIs C and D have 50\% time horizons $T_C$ and $T_D$, with $T_D \gg T_C$ while $T_B/T_A = T_D/T_C$. This represents the same multiplier of time horizon, and hence the same jump on the time-horizon plot, but it may not mean the same thing if the relationship between human time and AI difficulty changes significantly. 

Predictiveness has previously been recognized as an issue in general. When METR attempted to replicate its analysis on new task suites, they found low correlation between human time and AI success in the case of video understanding tasks \citep{HowDoesTime2025}, and diagnosed a lack of ``soundness of the time horizon metric'' in that setting. Later, \citet{mertensCrashingWavesVs2026a} found low predictiveness between human time and AI success on a broad range of economically relevant tasks. However, just because predictiveness is low does not mean 1-dimensional difficulties are not useful, and comparability does not seem to previously have been discussed. 

On METR's software tasks suite, predictiveness is strong. Using diagnostic plots, in this work we have found that it is \emph{locally} weak in the 2--30 min region, raising a comparability issue with other regions. Below we illustrate the use of our two diagnostic plots, the time-to-difficulty conversion and the conditional success plots, to assess both predictiveness and comparability, and hence calibrate our interpretations of the time horizons. 

\subsection{The time-to-difficulty conversion plot}\label{sec:diagnosedifficulty}

This section describes how we plot and interpret Figure \ref{fig:headline}(d). 

\paragraph{Plotting.} Under Model 2, there exists a parameter $\theta_i$ which represents the latent difficulty of the task, and which can be estimated under the context of the model. Here is how we estimated them using Model 2, but in principle they could be estimated with any IRT method. 

Let $x_i = \log_2 T_i$, and let $\widehat{\mathcal M}$ denote Model 2 with its fitted parameters. After fitting the model, we condition its joint distribution on all of the observed outcomes and compute
\[
    \hat\theta_i
    := \mathbb{E}_{\widehat{\mathcal M}}
       \left[\theta_i \mid \mathbf{Y}, \mathbf{T}\right].
\]
The blue point at $T_i$ is this posterior mean for task $i$. Thus, the point combines the evidence from every AI that attempted the task with the model's prior for a task of that human time.

Next, in Model 2, the time-to-difficulty conversion function $f$ is 
\[
    \mathbb{E}[\theta_i \mid \log_2 T_i = x] = f(x),
\]
where the expectation averages over both the mean-zero family effect and the task-specific residual. The purple band is $f(x) \pm s(x)$, where $s(x)^2 = \sigma(x)^2 + \tau(x)^2$ is the fitted variance for a new task from a new family at $\log_2$ human time $x$. 

The latent difficulty scale is identifiable only up to a positive affine transformation. For Figure \ref{fig:headline}(d), we therefore re-express it in terms of standard deviations. The dashed line plots $\text{SNR} = 1$: a straight line through the mean task that rises $1/\sqrt{2} \approx 0.71$ standard deviations of difficulty per standard deviation of log human time. 

\paragraph{Interpretation.}
To assess predictiveness from this chart, we simply ask whether the fitted shape is comparable to or steeper than the $\text{SNR} = 1$ line. If there is no slope, but nevertheless a large spread in difficulties, it could be either that the task difficulties are difficult to predict, or that we have not found the right predictor.  

To assess comparability, we can look directly at the fitted spline. Its changing shape, and the shape of the surrounding point cloud, suggest that time horizon jumps in the 2--30 min region should be interpreted differently from elsewhere. 

If the $\theta$'s are fit with IRT, these plots can also be used in exploratory data analysis to check what a reasonable functional form for $f$ might be, or to check against various proposed annotations. 

\subsection{Conditional success trajectories}\label{sec:th-vs-success}

This section describes how we plot and interpret Figure \ref{fig:headline}(b) and (f). 

\paragraph{Plotting.} For each fixed human time $t^* \in \{15\text{ sec}, 1\text{ min}, 4\text{ min}, 15\text{ min}, 1\text{ hr}, 4\text{ hr}, 16\text{ hr}\}$, the plot shows
\[
    p_j(t^*)
    = \Pr(Y_{ijr}=1 \mid T_i=t^*,\ \AI_j)
\]
on a logit scale against the AI's release date $D_j$, using the model predicted probabilities. 

\paragraph{Interpretation.}
The main fact to use in interpreting these plots is that the \emph{spread} between curves reflects predictiveness---the more spread out the curves are, the more predictive the annotation is---whereas any \emph{variation between spreads} suggests a comparability issue. 

In Figure \ref{fig:headline}(b), the model has reasonably spread out curves, suggesting good predictiveness, but the underlying baseline model used is not able to assess comparability due to its functional form. 

Figure \ref{fig:headline}(f) fits the data better and reveals how we should be adjusting for comparability. Reading time-horizon jumps by scanning the 0.5 line from left to right, the difference between the AI with a 4 min time horizon and the one with a 15 min time horizon is much smaller in terms of \emph{every} conditional probability compared to other jumps. 

Of course, aside from assessing construct validity of time horizons, the figure can also be used on its own merits to track capabilities. For this use, it is more difficult to read off that capabilities are advancing exponentially in some sense, and the logit scale is relatively unintuitive. Additionally, when multiple curves bunch up together and are not well-separated, such as the 4 and 15 min curves, the human time summary of the tasks is insufficient, when inspecting those curves, to interpret what ``sort'' of tasks the AI is succeeding at.

\subsection{Rasch abilities and the METR plot}

Finally, observe that the shared-slope logistic regression, Model 1, and Model 2 each estimate a quantity that directly targets the construct ``AI capability'', namely $\alpha_1, \dots, \alpha_J$ for $\AI_1, \dots, \AI_J$. When we are concerned that time horizons do not have construct validity, this could be plotted instead. We refer to them as Rasch abilities, in reference to the IRT Rasch model where a similar notion can be defined. 

\begin{figure}[ht]
    \centering
    \includegraphics[width=\textwidth]{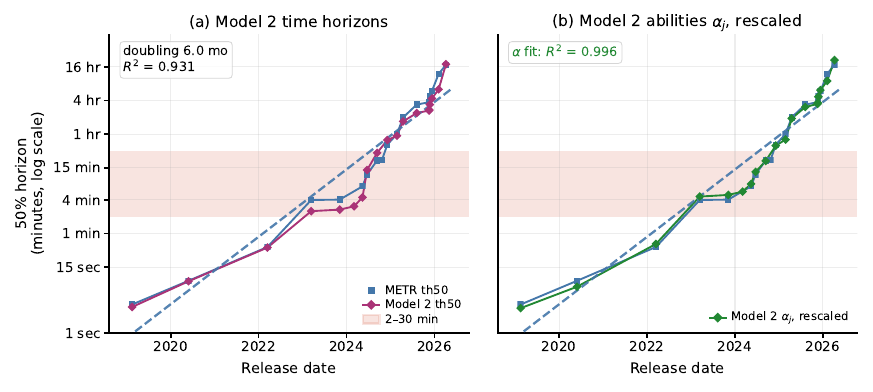}
    \caption{\textbf{Model 2 over the METR plot.} Both panels repeat METR's own per-AI time horizons (blue, with the trend dashed), which are close to the baseline horizons of Figure~\ref{fig:headline}(a). As in Figure~\ref{fig:headline}, only frontier AIs are shown, each series on its own frontier. \textbf{(a)} Model 2's time horizons (purple), the estimates of Figure~\ref{fig:headline}(e); the disagreement concentrates in the shaded 2--30 minute band. \textbf{(b)} Model 2's Rasch abilities $\hat\alpha_j$ (green), displayed as $2^{u + v\hat\alpha_j}$ minutes with $(u, v)$ chosen by least squares against $\log_2$ of the METR time horizons ($R^2 = 0.996$).}
    \label{fig:alpha-overlay}
\end{figure}

Figure~\ref{fig:alpha-overlay}(b) shows the results for Model 2, after an affine rescaling to compare it to the METR plot of Figure~\ref{fig:headline}(a); Figure~\ref{fig:alpha-overlay}(a) overlays Model 2's own time horizons on the same series, for reference. Strikingly, the rescaled abilities are \emph{almost identical} to the METR estimates---visibly closer than Model 2's own time horizons in panel (a). This means that even though the linear logistic model attempts to estimate time horizons, it may be inadvertently estimating another AI capability construct.

Why does this coincidence occur? To be clear, it is not guaranteed to occur, and would not have occurred if the shape of the spline in Figure~\ref{fig:headline}(d) diverged further from linear. In this case, note that under Model 2, the 50\% log-time horizon can be defined as $f^{-1}(\alpha_j / \beta_j)$ where $\beta_1 = \dots = \beta_J = \beta$, while in the linear model the log horizon is $\alpha_j / \beta_j$. So by plotting Model 2's results without $f^{-1}$, the coincidence is that, up to rescaling, the estimates $\alpha_j/\beta_j$ in each model are close to each other.

Because the flat region of $f$ is relatively localized to tasks within 2--30 min, while the estimation of $\alpha_j$ and $\beta$ depends on $\AI_j$'s performance on \emph{every} task, we believe that the linearity assumption on $f$ was close enough to correct on the whole task suite so that estimating $\alpha_j/\beta$ still gave a good answer. However, if we believe that $f$ is truly not linear, we should not then call the estimate a \emph{time horizon}, though we may still believe the measurements are nevertheless valuable as capturing \emph{Rasch ability}.

\section{Discussion: the future of time horizons}\label{sec:discussion}

\begin{figure}[t]
    \centering
    \includegraphics[width=0.6\textwidth]{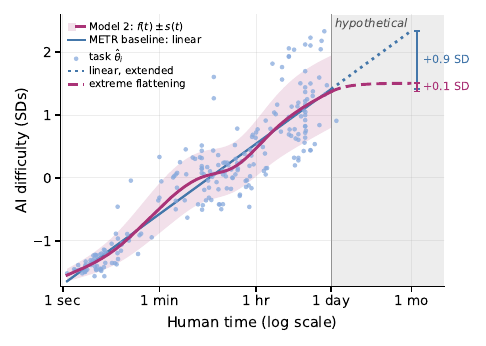}
    \caption{\textbf{A hypothetical time-to-difficulty conversion beyond one day.} Up to one day, the plot is Figure~\ref{fig:headline}(d); after that, we depict a possible future for the time-to-difficulty conversion.}\label{fig:future-conversion}
\end{figure}

It seems difficult to extend the time-horizon benchmark beyond the human time length of tasks that is currently available (up to 30 hours). However, suppose it were possible. Then there is no a priori reason that newly constructed, longer tasks should preserve the approximately linear relationship observed over the current range; future portions of $f$ may again be unusually flat or steep. It seems quite plausible that the estimated purple curve (and perhaps an underlying true pattern) could look like Figure \ref{fig:future-conversion}, and checks on a linear model will not detect this, while our diagnostic plots will. 

Additionally, when assessing AI's effects in the broader economy outside of software engineering, capability anchoring seems of great importance for scientific communication of the speed and risks of development. One domain with great opportunity is in physical intelligence, i.e. robotics. In this case, it seems of great importance to pick a good annotation to carefully assess the construct validity of time horizons to communicate advancement to the public. However, the best task annotation may not be human time. 

\begin{ack}
    We thank Alexander Barry, Miles Tidmarsh, Jonathan Gabor, and the AI safety community for useful discussions. 
\end{ack}

\bibliography{METRbumps.bib}

\clearpage
\appendix

\section{Appendix}

\subsection{Per-AI evaluations}\label{app:perai}

\begin{figure}[h]
    \centering
    \includegraphics[width=\textwidth]{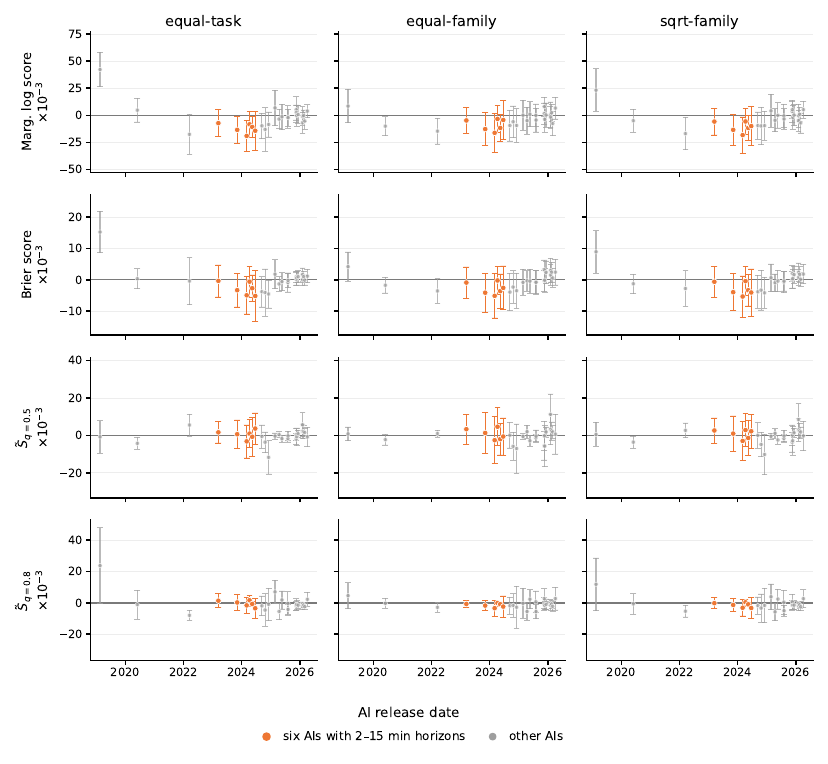}
    \caption{\textbf{Per-AI scoring-rule contrasts, Model 1 $-$ baseline}, for all 26 AIs in release order under family CV. Each point is the paired difference on that AI's held-out cells ($\times 10^{-3}$; below zero favors Model 1), with a 95\% Bayle cluster-robust interval using the task families that AI attempted as clusters. Weights and family sizes are renormalized within each AI. Columns give the three evaluation weightings and rows the marginal log score, Brier score, and the smoothed elementary scores $\tilde S_{q=0.5}$ and $\tilde S_{q=0.8}$, which average over $q \in [0.45,0.55]$ and $q \in [0.75,0.85]$, respectively. Orange marks the six AIs of Figure~\ref{fig:pfits}, whose METR time horizons are 2--15 min; all other AIs are gray.}
    \label{fig:perai-gain-m1}
\end{figure}

\clearpage
\begin{figure}[ht]
    \vspace*{0.5in}
    \centering
    \includegraphics[width=\textwidth]{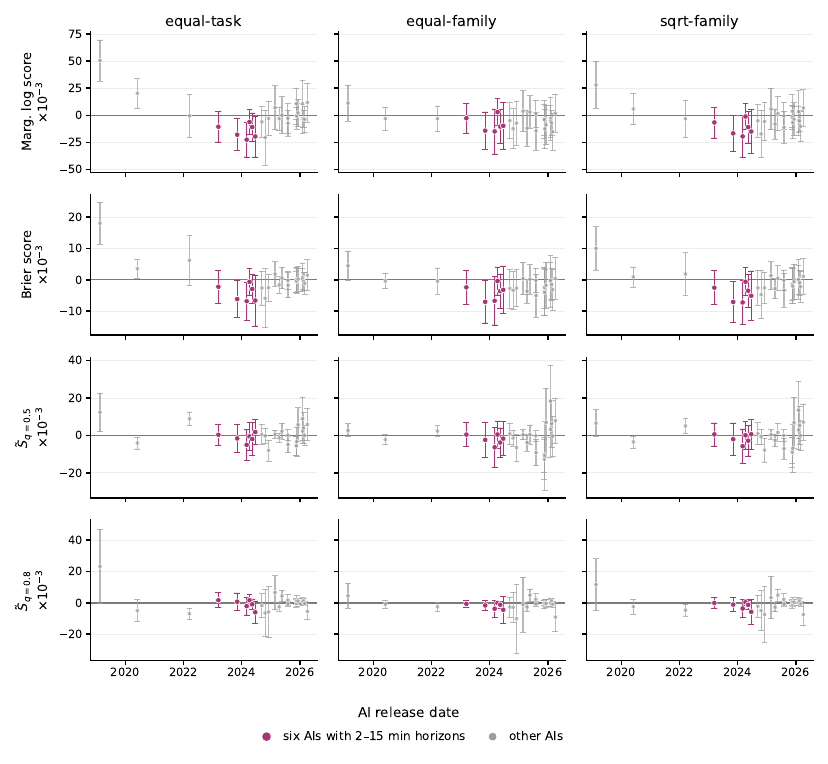}
    \caption{\textbf{Per-AI scoring-rule contrasts, Model 2 $-$ baseline}, with the same construction and shared score-wise axis limits as Figure~\ref{fig:perai-gain-m1}. Purple marks the same six AIs with time horizons of 2--15 min. The point-estimate advantage is concentrated in the middle of the capability range, while most individual-AI intervals remain too wide to separate from zero; the pooled comparison in Figure~\ref{fig:scoregrid} is where the evidence accumulates.}
    \label{fig:perai-gain-m2}
\end{figure}

\clearpage
\begin{figure}[ht]
    \centering
    \includegraphics[height=0.85\textheight,keepaspectratio]{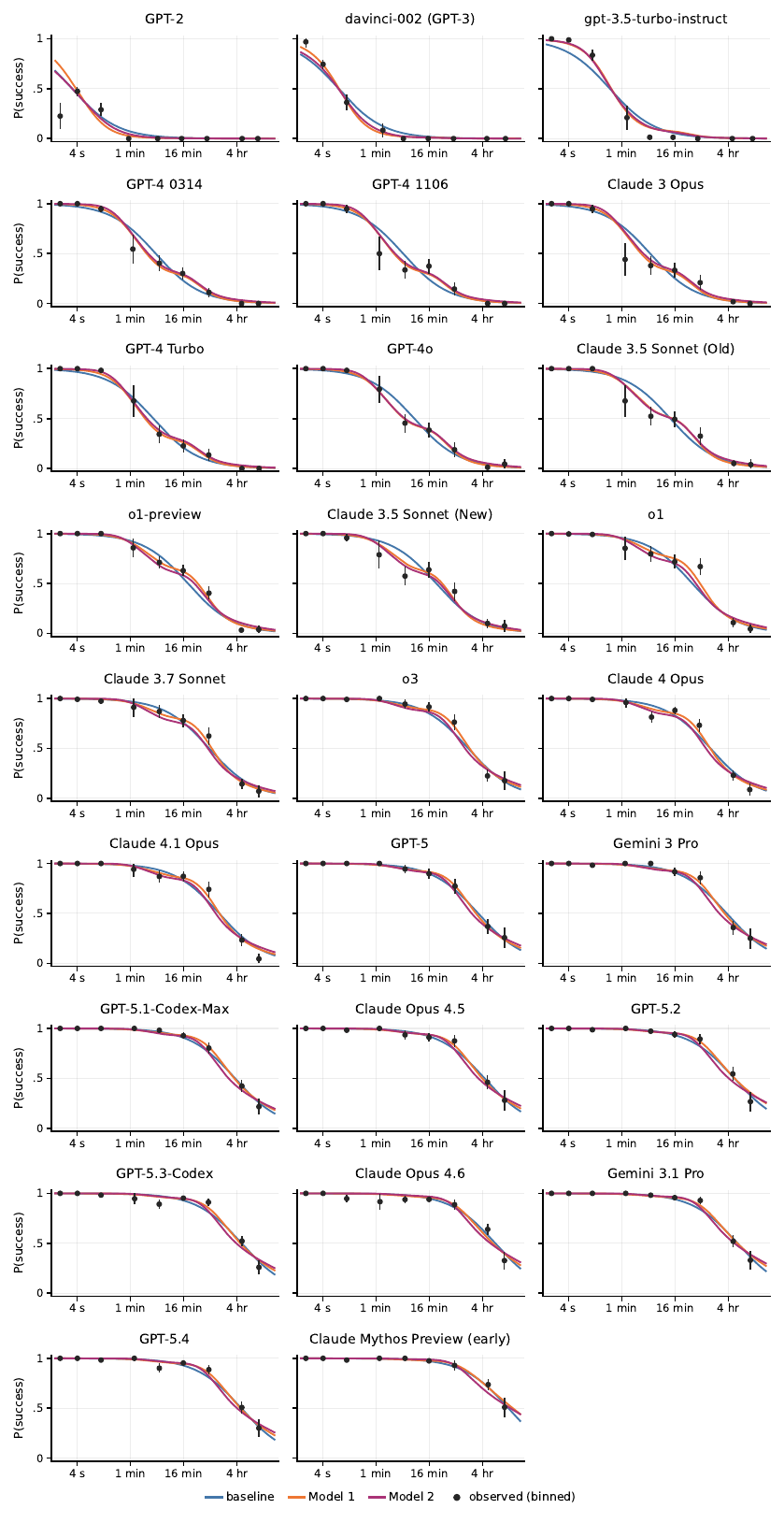}
    \caption{\textbf{In-sample fitted $p_j(t)$ against binned observed frequencies for all 26 AIs}, in release order; construction as in Figure~\ref{fig:pfits} (factor-4 human-time bins centered on 1 min, 4 min, 16 min, \dots; points at run-weighted mean human times, 95\% Wald intervals treating runs as independent). The shape misfit of the log-linear baseline is concentrated in the mid-strength AIs of the GPT-4 era; for the weakest AIs the curves differ mainly in the tail, and for the strongest the three specifications nearly coincide.}
    \label{fig:pfits-all}
\end{figure}

\end{document}